\pdfoutput=1
\documentclass[11pt]{article}

\usepackage{ACL2023}
\usepackage{times}
\usepackage{latexsym}
\usepackage[T1]{fontenc}
\usepackage[utf8]{inputenc}
\usepackage{times}
\usepackage{latexsym}
\usepackage{bm}
\usepackage{algorithm}
\usepackage{algorithmic}
\usepackage{multirow}
\usepackage{amsfonts}
\usepackage{color}
\usepackage{amsmath, mathtools}
\usepackage{microtype}

\usepackage{inconsolata}

\title{Is ChatGPT A Good Keyphrase Generator? A Preliminary Study}

\author{Mingyang Song$^{\spadesuit}$, Haiyun Jiang$^{\clubsuit}$$^\ast$, Shuming Shi$^{\clubsuit}$\\
	{\bf Songfang Yao$^{\spadesuit}$, Shilong Lu$^{\spadesuit}$, Yi Feng$^{\spadesuit}$, Huafeng Liu$^{\spadesuit}$, Liping Jing$^{\spadesuit}$}\thanks{\ \ Corresponding Author} \\
	$^{\clubsuit}$Tencent AI Lab, Shenzhen, China \\
	$^{\spadesuit}$Beijing Key Lab of Traffic Data Analysis and Mining \\
	$^{\spadesuit}$Beijing Jiaotong University, Beijing, China \\
	{\tt mingyang.song@bjtu.edu.cn} \\
}

\begin{document}
\maketitle
\begin{abstract}
%Recently, the emergence of ChatGPT has attracted wide attention from the computational linguistics community. Therefore, in this report, we provide a preliminary evaluation of ChatGPT for the keyphrase generation task to show its capability as a keyphrase generator, such as keyphrase generation prompts, keyphrase generation diversity, multi-domain keyphrase generation, and long document understanding. We adopt the prompt advised by OpenAI and extend it to six candidate prompts to trigger the keyphrase generation ability of ChatGPT and find that the candidate prompts generally work well and show minor performance differences on six benchmark datasets. Overall, we find that ChatGPT has excellent potential for keyphrase generation, and we also list several limitations and future expansions of this report in the final section. In addition, we find that absent keyphrase generation is still a challenge for ChatGPT. 
The emergence of ChatGPT has recently garnered significant attention from the computational linguistics community. To demonstrate its capabilities as a keyphrase generator, we conduct a preliminary evaluation of ChatGPT for the keyphrase generation task. We evaluate its performance in various aspects, including keyphrase generation prompts, and keyphrase generation diversity. Our evaluation is based on six benchmark datasets, and we adopt the prompt suggested by OpenAI while extending it to six candidate prompts. 
Overall, we find that ChatGPT performs exceptionally well on all six candidate prompts, with minor performance differences observed across the datasets. Based on our findings, we conclude that ChatGPT has great potential for keyphrase generation. Moreover, we discover that ChatGPT still faces challenges when it comes to generating absent keyphrases. Meanwhile, in the final section, we also present some limitations and future expansions of this report. Furthermore, we have released the related data\footnote{\url{https://github.com/MySong7NLPer/ChatGPT_as_Keyphrase_Generator}}.

%\footnote{{The used data will be released in the future.}}.

\end{abstract}

\section{Introduction}
ChatGPT\footnote{\url{https://openai.com/blog/chatgpt}} is an intelligent chatbot developed by OpenAI based on the InstructGPT \cite{instructgpt}. ChatGPT aims to provide excellent conversation responses based on the prompts provided. OpenAI highlights that ChatGPT can provide detailed responses, answer follow-up questions, and challenge incorrect or inappropriate prompts while maintaining a natural dialogue format. Furthermore, ChatGPT encompasses a wide range of natural language processing abilities, such as code debugging, question answering, article paraphrasing, keyphrase generation, and logic reasoning, to name a few. We are particularly intrigued by the capability of keyphrase generation of ChatGPT, as we would like to understand how well it performs compared to the state-of-the-art keyphrase generation models. We focus on narrowing the gap between ChatGPT and such keyphrase generation models and exploring new and innovative ways to improve the performance of keyphrase generation and enhance its features to provide even better results. Note that we consider that keyphrases and keywords are different in this report. Keyphrases are often phrases of two or more words rather than single words (keywords). %Furthermore, we present corresponding analysis on this issue in the report.

\begin{figure}
	\centering
	\includegraphics[scale=0.5]{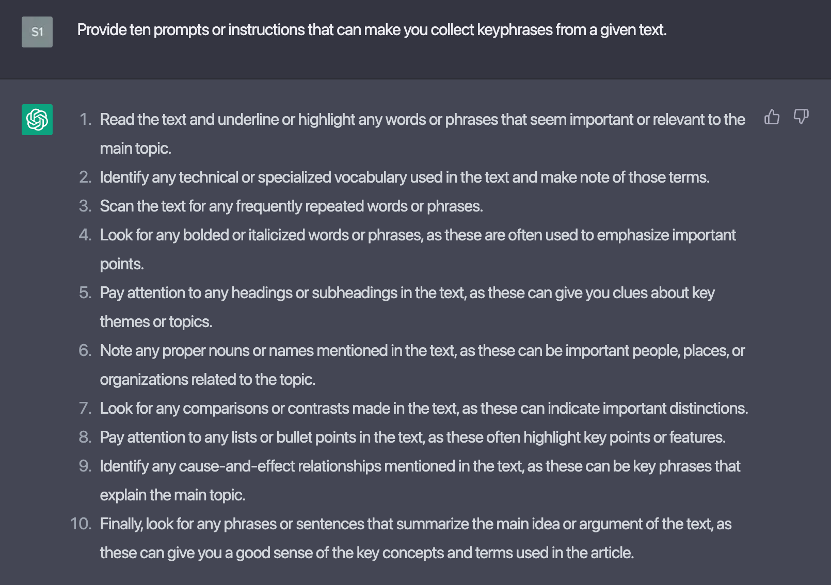}
	\caption{Prompts or instructions advised by ChatGPT for keyphrase generation.}% (Date: 2023.03.01).}
	\label{prompts}
\end{figure}

\begin{table*}[t!]
	\begin{center}
		\scriptsize
		\renewcommand\arraystretch{1.7}
		\renewcommand\tabcolsep{11pt}
		\begin{tabular}{l|c|c|c|c|c}
			\hline \hline 
			\textbf{\textsc{Test Set}} & \textbf{\textsc{Domain}}  & \textbf{\textsc{Type}} & \textbf{\textsc{\# Doc.}} & \textbf{\textsc{Avg. \# Words}} & \textbf{Present KPs (\%)} \\ \hline
			\textsc{KP20k} \cite{catseq17}  &  Scientific Abstract  & Short & {20,000} & 179.8 & 57.40 \\
			\textsc{Inspec} \cite{Inspec} &  Scientific Abstract & Short & {500} & 128.7 & 55.69 \\
			\textsc{Nus} \cite{Nus} &  Scientific Abstract & Short & {211} & 219.1 & 67.75 \\
			\textsc{Krapivin} \cite{Krapivin} &  Scientific Abstract & Short & 400 & 182.6 & 44.74  \\
			\textsc{SemEval2010} \cite{SemEval} &  Scientific Abstract & Short & {100} & 234.8 & 42.01 \\
			%\textsc{DUC2001} \cite{duc2001_singlerank} &  News Article & Medium & {\color{red}\textbf{308}} & 847.2 & 97.82  \\
			%\textsc{OpenKP} \cite{xiong19} &  Open Web Domain & Long & {\color{red}\textbf{6,616}} & 900.4 & 100.00   \\
			%\textsc{SemEval2010} \cite{SemEval} &  Full Scientific Paper & Long & {\color{red}\textbf{100}} & 1587.5 & 88.70   \\
			\hline\hline
		\end{tabular}
	\end{center}
	\caption{\label{dataset} Information of adopted test sets. {\# Doc.} is the number of documents in the dataset.  {Avg. \# Words} is the average number of words for documents. {Present KPs (\%)} indicates the percentage of keyphrases, which are presented in the documents. {Note that this report uses all of the test data rather than sampling part from it}.}
\end{table*}

In this report, we summarize a preliminary study of ChatGPT on keyphrase generation to gain a better understanding of it. Specifically, we focus on five perspectives:

\noindent{\textbf{$\bullet$ Keyphrase Generation Prompts}:} 

ChatGPT is a large language model that relies on prompts to guide its keyphrase generation. The choice of prompts or instructions can significantly impact the quality of the generated keyphrases.
In general, the performance of ChatGPT in generating keyphrases relies heavily on the quality of the prompts. More specifically, a well-crafted prompt yields better results than a poorly formulated one. As a result, we assess the effectiveness of six different prompts to evaluate ChatGPT's performance in the task of keyphrase generation.
%\noindent{\textbf{$\bullet$ Keyphrase Generation Robustness}:} 

\noindent{\textbf{$\bullet$ Keyphrase Generation Diversity}:} 

The goal of keyphrase generation is to produce a collection of phrases that encompass the main topics discussed in a given document, as highlighted in \cite{2014survey, song_survey}. Recent advancements in keyphrase generation have demonstrated remarkable progress, as evidenced by improved quality metrics like the F1 score. However, the importance of diversity in keyphrase generation has often been overlooked, as discussed in \cite{diversity}. In line with previous studies \cite{one2set,wrone2set,setmatch, diversity}, we present statistics on the average numbers of unique present and absent keyphrases and the average duplication ratios of all generated keyphrases. These metrics allow us to measure the ability of ChatGPT to generate diverse keyphrases.

%Keyphrase generation aims to generate a set of phrases that can cover the main topics discussed in a given document \cite{2014survey}. Recent advances in keyphrase generation have made remarkable progress, demonstrated through improved quality metrics such as F1-score. However, the importance of diversity in keyphrase generation has been largely ignored \cite{diversity}. Following previous studies \cite{one2set, wrone2set}, we report the average numbers of unique present and absent keyphrases and the average duplication ratios of all predicted keyphrases to investigate the ability of ChatGPT to generate diverse keyphrases.

%\noindent{\textbf{$\bullet$ Multi-domain Keyphrase Generation}:} 

%ChatGPT is a versatile language model that performs various natural language processing tasks across multiple domains. We are interested in evaluating how well ChatGPT performs on different domains, considering differences in text styles (e.g., news documents vs. scientific articles). This will give us a better understanding of its strengths and limitations in different contexts.

%\footnotetext[2]{\url{https://platform.openai.com/examples/default-keywords}}

\begin{table}[t!]
	\renewcommand\arraystretch{1.5}
	\begin{center}
		\scriptsize
		\begin{tabular}{p{0.5cm}p{6cm}}
			\hline\hline
			
			\multicolumn{2}{c}{\textsc{Prompts}} \\ \hline

			\textsc{\textbf{Tp}$1^\dagger$}    & \textbf{\texttt{Extract keywords from this text:}} \texttt{[\textsc{Document}]} \\ \hline
			\textsc{\textbf{Tp}$2$}   & \texttt{Generate keywords from this text:} \texttt{[\textsc{Document}]} \\ 
			\textsc{\textbf{Tp}$3$}    & \texttt{Extract keyphrases from this text:} \texttt{[\textsc{Document}]} \\ 
			\textsc{\textbf{Tp}$4$}    & \texttt{Generate keyphrases from this text:} \texttt{[\textsc{Document}]} \\ 
			\textsc{\textbf{Tp}$5$}    & \texttt{Generate present and absent keywords from this text:} \texttt{[\textsc{Document}]} \\ 
			\textsc{\textbf{Tp}$6$}    & \texttt{Generate present and absent keyphrases from this text:} \texttt{[\textsc{Document}]} \\ 
			\hline\hline
			
		\end{tabular}
		\caption{\label{case} Six prompts are designed for chatting with ChatGPT to collect keyphrases from the given text document. $^\dagger$ indicates the prompt provided by OpenAI.}%\footnotemark[2].
	\end{center}
\end{table}

\begin{table*}[!htb]
	\scriptsize
	\centering
	\renewcommand\tabcolsep{7pt}
	\renewcommand\arraystretch{1.4}
	\begin{tabular}{c|cc|cc|cc|cc|cc}
		\hline\hline
		\multirow{2}{*}{ \textbf{\textsc{Model}}} & \multicolumn{2}{c|}{\textsc{\textbf{KP20k}}} & \multicolumn{2}{c|}{\textsc{\textbf{Inspec}}}& \multicolumn{2}{c|}{\textsc{\textbf{Nus}}}& \multicolumn{2}{c|}{\textsc{\textbf{Krapivin}}}& \multicolumn{2}{c}{\textsc{\textbf{SemEval}}}\\ %\cline{2-2} 
		& F1@5 & F1@M & F1@5 & F1@M & F1@5 & F1@M& F1@5 & F1@M& F1@5 & F1@M \\ \hline
		\multicolumn{11}{l}{\textsc{RNN-based Models}} \\\hline
		\multicolumn{1}{l|}{\textsc{catSeq} \cite{catseq17}}
		& 0.291 & 0.367 & 0.225 & 0.262 & 0.323 & 0.397 & 0.269 & 0.354 & 0.242 & 0.283 \\ 
		\multicolumn{1}{l|}{\textsc{catSeqTG-2RF1} \cite{adaptive_reward}}
		& 0.321 & 0.386 & 0.253 & 0.301 & 0.375 & 0.433 & 0.300 & 0.369 & 0.287 & 0.329 \\ 
		\multicolumn{1}{l|}{\textsc{ExHiRD-h} \cite{ExHiRD}}
		& 0.311 & 0.374 & 0.253 & 0.291 & N/A & N/A & 0.286 & 0.347 & 0.284 & 0.335 \\ \hline\hline

		\multicolumn{11}{l}{\textsc{Transformer-based Models}} \\\hline
		\multicolumn{1}{l|}{{\textsc{SetTrans}} \cite{one2set}}
		& 0.358 & 0.392 & 0.285 & 0.324 & 0.406 & 0.450 & 0.326 & \textbf{0.364} & 0.331 & 0.357 \\ 
		\multicolumn{1}{l|}{{\textsc{WR-SetTrans}} \cite{wrone2set}}
		& \textbf{0.370} & \textbf{0.378} & {0.330} & {0.351} & \textbf{0.428} & \textbf{0.452} & \textbf{0.360} & {0.362} & \textbf{0.360} & \textbf{0.370} \\\hline\hline
		
		\multicolumn{11}{l}{\textsc{PLM-based Models}} \\\hline
		
		\multicolumn{1}{l|}{\textsc{SciBART-base$^\ddagger$ (124M)}}
		& 0.341 & 0.396 & 0.275 & 0.328 & 0.373 & 0.421 & 0.282 & 0.329 & 0.270 & 0.304 \\ 
		\multicolumn{1}{l|}{\textsc{BART-base$^\ddagger$ (140M)}}
		& 0.322 & 0.388 & 0.270 & 0.323 & 0.366 & 0.424 & 0.270 & 0.336 & 0.271 & 0.321 \\ 
		\multicolumn{1}{l|}{\textsc{T5-base$^\ddagger$ (223M)}}
		& 0.336 & 0.388 & 0.288 & 0.339 & 0.388 & 0.440 & 0.302 & 0.350 & 0.295 & 0.326 \\ 
		\hline\hline
		
		\multicolumn{11}{l}{\textsc{ChatGPT\ (gpt-3.5-turbo)}} \\\hline
		\multicolumn{1}{l|}{\textsc{ChatGPT w/ \textbf{Tp}$1^\dagger$}}
		& 0.186 & 0.160 & 0.298 & 0.417 & 0.319 & 0.225 & 0.239 & 0.187 & 0.267 & 0.216 \\
		\multicolumn{1}{l|}{\textsc{ChatGPT w/ \textbf{Tp}$2$}}
		& 0.180 & 0.149 & 0.310 & 0.433 & 0.314 & 0.239 & 0.243 & 0.197 & 0.275 & 0.240 \\
		\multicolumn{1}{l|}{\textsc{ChatGPT w/ \textbf{Tp}$3$}}
		& 0.161 & 0.141 & 0.383 & 0.463 & 0.281 & 0.211 & 0.218 & 0.184 & 0.268 & 0.214 \\
		\multicolumn{1}{l|}{\textsc{ChatGPT w/ \textbf{Tp}$4$}}
		& 0.160 & 0.136 & 0.322 & 0.393 & 0.208 & 0.187 & 0.170 & 0.163 & 0.233 & 0.212 \\
		\multicolumn{1}{l|}{\textsc{ChatGPT w/ \textbf{Tp}$5$}}
		& 0.174 & 0.161 & 0.330 & 0.427 & 0.336 & 0.261 & 0.244 & 0.204 & 0.281 & 0.243 \\
		\multicolumn{1}{l|}{\textsc{ChatGPT w/ \textbf{Tp}$6$}}
		& 0.179 & 0.165 & \textbf{0.401} & \textbf{0.476} & 0.287 & 0.253 & 0.230 & 0.213 & 0.262 & 0.233 \\
		
		\hline\hline
	\end{tabular}
	
	\caption{Results of present keyphrase generation on five datasets. F1 scores on the top 5 and M keyphrases are reported, where M is a variable cut-off equal to the number of predictions. The results of the baselines are reported in their corresponding papers. Specifically, $^\ddagger$ indicates the results are reported in \citet{wudi}. The best results are highlighted in bold.}
	\label{present}
\end{table*}
\section{ChatGPT for Keyphrase Generation}
\subsection{Evaluation Setting}
We briefly introduce the evaluation setting, which mainly includes the compared baselines, datasets, and evaluation metrics.
Note that each time a new query is made to ChatGPT, we clear conversations to avoid the influence of previous samples, which is similar to \citet{gangda}.

%\subsubsection{Baselines}
We compare ChatGPT with several state-of-the-art keyphrase generation systems: \textsc{catSeq} \cite{catseq17}, \textsc{catSeqTG-2RF1} \cite{adaptive_reward}, \textsc{ExHiRD-h} \cite{ExHiRD}, \textsc{SetTrans} \cite{one2set}, and \textsc{WR-SetTrans} \cite{wrone2set}. By default, the results in this report come from the ChatGPT version on 2023.03.01. For new results, we will mark the updated version information correspondingly.

%\subsubsection{Datasets}
We evaluate ChatGPT and all the baselines on five datasets: \textsc{Inspec} \cite{Inspec}, \textsc{NUS} \cite{Nus}, \textsc{Krapivin} \cite{Krapivin}, \textsc{SemEval} \cite{SemEval}, and \textsc{KP20k} \cite{catseq17}. As implemented in \cite{one2set, wrone2set}, we perform data preprocessing, including tokenization, lowercasing, replacing all digits with the symbol ⟨digit⟩, and removing duplicated instances. Note that all the baselines were trained by using the KP20k training set. Furthermore, we verify the keyphrase generation ability of ChatGPT in real-world scenarios on the \textsc{OpenKP} dataset \cite{xiong19} where \textit{documents are from diverse domains} and have variant content quality. Table~\ref{dataset} summarizes the information of the used dataset.
\begin{table*}[!htb]
	\scriptsize
	\centering
	\renewcommand\tabcolsep{7pt}
	\renewcommand\arraystretch{1.4}
	\begin{tabular}{c|cc|cc|cc|cc|cc}
		\hline\hline
		\multirow{2}{*}{ \textsc{\textbf{{Model}}}} & \multicolumn{2}{c|}{\textsc{\textbf{KP20k}}} & \multicolumn{2}{c|}{\textsc{\textbf{Inspec}}}& \multicolumn{2}{c|}{\textsc{\textbf{Nus}}}& \multicolumn{2}{c|}{\textsc{\textbf{Krapivin}}}& \multicolumn{2}{c}{\textsc{\textbf{SemEval}}}\\ %\cline{2-2} 
		& F1@5 & F1@M & F1@5 & F1@M & F1@5 & F1@M& F1@5 & F1@M& F1@5 & F1@M \\ \hline
		
		\multicolumn{11}{l}{\textsc{RNN-based Models}} \\\hline
		\multicolumn{1}{l|}{\textsc{catSeq} \cite{catseq17}}
		& 0.015 & 0.032 & 0.004 & 0.008 & 0.016 & 0.028 & 0.018 & 0.036 & 0.016 & 0.028 \\ 
		\multicolumn{1}{l|}{\textsc{catSeqTG-2RF1} \cite{adaptive_reward}}
		& 0.027 & 0.050 & 0.012 & 0.021 & 0.019 & 0.031 & 0.030 & 0.053 & 0.021 & 0.030 \\ 
		\multicolumn{1}{l|}{\textsc{ExHiRD-h} \cite{ExHiRD}}
		& 0.016 & 0.032 & 0.011 & 0.022 & N/A & N/A & 0.022 & 0.043 & 0.017 & 0.025 \\ 
		\hline\hline
		\multicolumn{11}{l}{\textsc{Transformer-based Models}} \\\hline
		\multicolumn{1}{l|}{{\textsc{SetTrans}} \cite{one2set}}
		& 0.036 & {0.058} & 0.021 & 0.034 & 0.042 & 0.060 & 0.047 & 0.073 & 0.026 & 0.034 \\ 
		\multicolumn{1}{l|}{{\textsc{WR-SetTrans}} \cite{wrone2set}}
		& \textbf{0.050} & \textbf{0.064} & 0.025 & 0.034 & \textbf{0.057} & \textbf{0.071} & \textbf{0.057} & \textbf{0.074} & \textbf{0.040} & \textbf{0.043} \\
		\hline\hline
		
		\multicolumn{11}{l}{\textsc{PLM-based Models}} \\\hline
		\multicolumn{1}{l|}{\textsc{SciBART-base$^\ddagger$ (124M)}}
		& 0.029 & 0.052 & 0.016 & 0.028 & 0.033 & 0.053 & 0.033 & 0.054 & 0.018 & 0.022 \\ 
		\multicolumn{1}{l|}{\textsc{BART-base$^\ddagger$ (140M)}}
		& 0.022 & 0.042 & 0.010 & 0.017 & 0.026 & 0.042 & 0.028 & 0.049  & 0.016 & 0.021 \\ 
		\multicolumn{1}{l|}{\textsc{T5-base$^\ddagger$ (223M)}}
		& 0.017 & 0.034 & 0.011 & 0.020 & 0.027 & 0.051 & 0.023 & 0.043 & 0.014 & 0.020 \\

		\hline\hline
		\multicolumn{11}{l}{\textsc{ChatGPT\ (gpt-3.5-turbo)}} \\\hline
		\multicolumn{1}{l|}{\textsc{ChatGPT w/ \textbf{T\scriptsize{P}}$1^\dagger$}}
		& 0.045 & 0.053 & 0.016 & 0.030 & 0.001 & 0.001 & 0.003 & 0.004 & 0.006 & 0.007 \\
		\multicolumn{1}{l|}{\textsc{ChatGPT w/ \textbf{Tp}$2$}}
		& 0.045 & 0.052 & 0.025 & 0.042 & 0.008 & 0.009& 0.007 & 0.011 & 0.005 & 0.007 \\
		\multicolumn{1}{l|}{\textsc{ChatGPT w/ \textbf{Tp}$3$}}
		& 0.041 & 0.045 & 0.027 & 0.047 & 0.003 & 0.004 & 0.004 & 0.008 & 0.002 & 0.002 \\
		\multicolumn{1}{l|}{\textsc{ChatGPT w/ \textbf{Tp}$4$}}
		& 0.038 & 0.039 & \textbf{0.030} & \textbf{0.041} & 0.009 & 0.012 & 0.011 & 0.015 & 0.004 & 0.005 \\
		\multicolumn{1}{l|}{\textsc{ChatGPT w/ \textbf{Tp}$5$}}
		& 0.036 & 0.026 & 0.017 & 0.015 & 0.004 & 0.005 & 0.006 & 0.005 & 0.004 & 0.004 \\
		\multicolumn{1}{l|}{\textsc{ChatGPT w/ \textbf{Tp}$6$}}
		& 0.041 & 0.025 & 0.029 & 0.024 & 0.009 & 0.009 & 0.005 & 0.009 & 0.005 & 0.006 \\
		
		\hline\hline
	\end{tabular}
	
	\caption{Results of absent keyphrase generation on five datasets. F1 scores on the top 5 and M keyphrases are reported, where M is a variable cut-off equal to the number of predictions. The results of the baselines are reported in their corresponding papers. Specifically, $^\ddagger$ indicates the results are reported in \citet{wudi}.}% The best results are highlighted in bold.}
	\label{absent}
\end{table*}

%\subsubsection{Evaluation Metrics}
Following previous studies \cite{catseq17}, we adopt macro averaged F1@5 and F1@M to evaluate the quality of both present and absent keyphrases. When using F1@5, blank keyphrases are added to make the keyphrase number reach five if the prediction number is less than five. Similar to the previous work \cite{kiemp, hypermatch, csl, icnlsp, hguke, setmatch, diversityrank}, we employ the Porter Stemmer to remove the identical stemmed keyphrases.

\subsection{Keyphrase Generation Prompts}
To design the prompts or instructions for triggering the keyphrase generation ability of ChatGPT, we seek inspiration from ChatGPT by asking it for advice. Concretely, we ask ChatGPT with the following instruction:

{{$\bullet$ \ \ \texttt{\small Provide ten prompts or instructions that can make you collect keyphrases from a given text.}}} 

%\noindent and obtain the results in Figure~\ref{prompts}. 
%The generated prompts seem reasonable, as they all involve extracting some core features of the keyphrases. However, we found that using these prompts to ask ChatGPT, what we got was not a list of keyphrases but something like a summary. Fortunately, OpenAI provides an official prompt on how to extract keywords. Based on this instruction, we extend it to six different candidate prompts, as shown in Table~\ref{case}. Specifically, we expect to analyze ChatGPT's understanding of several of the following through these six prompts:
More specifically, the generated prompts appear reasonable, as they all pertain to extracting important features of keyphrases. However, when we utilized these prompts to query ChatGPT, the responses we received were not lists of keyphrases but something more akin to summaries. Fortunately, OpenAI offers an official prompt for extracting keyphrases. Building upon this instruction, we expanded it into six distinct candidate prompts, as illustrated in Table~\ref{case}. Our specific aim with these six prompts is to evaluate the comprehension of ChatGPT on the following aspects:
\begin{itemize}
	\item[1.] \textit{What is the difference between “keyword” and “keyphrase” ?}
	
	From the results in Table~\ref{present}, it seems that ChatGPT does not have a gap in the understanding between "keyword" and "keyphrase".
	
	\item[2.] \textit{What is the difference between “extract” and “generate” ?}
	
	As you can see from Table~\ref{dup}, the prompts using “generate” (e.g., $\textsc{\textbf{Tp}2}$ and $\textsc{\textbf{Tp}4}$) predict more absent keyphrases than the prompts using “extract” (e.g., $\textsc{\textbf{Tp}1$^\dagger$}$ and $\textsc{\textbf{Tp}3}$) when chatting with ChatGPT. This shows that ChatGPT makes a distinction between present and absent keyphrases.
	
	\item[3.] \textit{What is the difference between “present keyphrase” and “absent keyphrase” ?}
	
	Figure~\ref{present_absent} illustrates the answer of ChatGPT to this question.
	From the results, we find that ChatGPT distinguishes between present and absent keyphrases. When explicitly asked in the instruction to generate present and absent will significantly increase the number of generated absent keyphrases by ChatGPT. More importantly, we find that ChatGPT is really not good at generating absent keyphrases.% in this report.
	
	\begin{figure}
		\centering
		\includegraphics[scale=0.35]{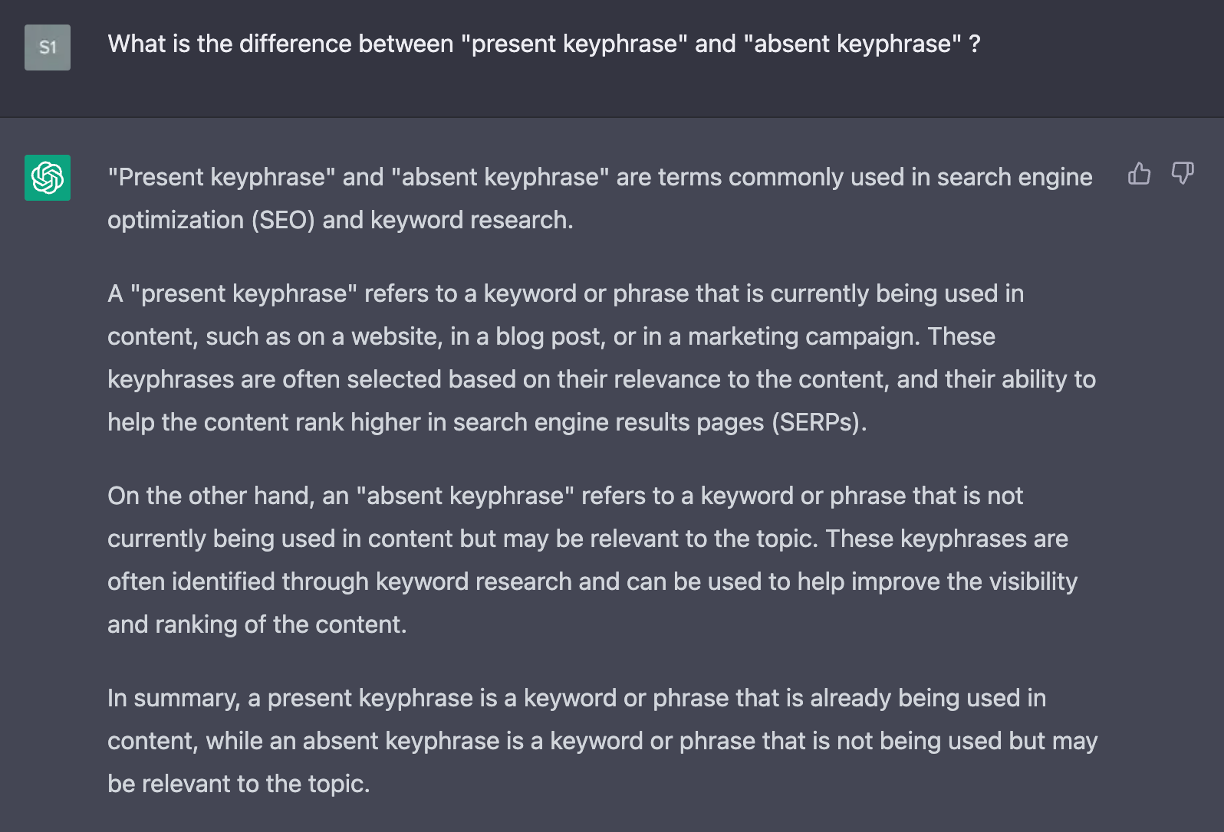}
		\caption{The answer of ChatGPT about the difference between “present keyphrase” and “absent keyphrase”.}
		\label{present_absent}
	\end{figure}
\end{itemize}

%\subsection{Multi-domain Keyphrase Generation}
%To verify the adaptability of ChatGPT under different domain documents, we tested it on the OpenKP dataset. The results show that ChatGPT is highly adaptable to multi-domain data and performs better than the state-of-the-art keyphrase extraction baselines. 

%\subsection{Keyphrase Generation Robustness}
%This report synthesizes the capability of ChatGPT in keyphrase generation with \textit{different prompts} as trigger to handle \textit{different domains} and \textit{documents of different lengths} to verify its robustness.

\begin{table}[t!]
	\begin{center}
		\scriptsize
		\renewcommand\arraystretch{1.4}
		\renewcommand\tabcolsep{4pt}
		\begin{tabular}{l|ccc|ccc}
			\hline \hline 
			
			\multirow{2}{*}{ \textbf{{Model}}} & \multicolumn{3}{c|}{\textsc{\textbf{KP20k}}} & \multicolumn{3}{c}{\textsc{\textbf{SemEval}}} \\ \cline{2-7}
			& \#PK & \#AK & Dup  & \#PK & \#AK & Dup  \\ \hline
			\textbf{\color{black}\textsc{Oracle}}  &  \textbf{\color{black}\textsc{3.31}}  & \textbf{\color{black}\textsc{1.95}} & \textbf{\color{black}\textsc{0.000}}  &  \textbf{\color{black}\textsc{6.12}}  & \textbf{\color{black}\textsc{8.31}} & \textbf{\color{black}\textsc{0.000}} \\\hline

			\multicolumn{1}{l|}{\textsc{SetTrans}}  &  \textbf{5.10}  & \textbf{2.01} & 0.080  &  4.62  & 2.18 & 0.080 \\
			\multicolumn{1}{l|}{\textsc{WR-SetTrans}}  &  6.35  & 3.26 & 0.100  &  \textbf{5.94}  & \textbf{3.60} & 0.100 \\\hline
			\multicolumn{1}{l|}{\textsc{ChatGPT w/ \textbf{Tp}$1^\dagger$}}
			& 14.14 & 5.58 & \textbf{0.005} & 24.53 & 1.06 & \textbf{0.009} \\
			\multicolumn{1}{l|}{\textsc{ChatGPT w/ \textbf{Tp}$2$}}
			& 16.29 & 6.33 & 0.006 & 21.78 & 1.66 & 0.013
			\\
			\multicolumn{1}{l|}{\textsc{ChatGPT w/ \textbf{Tp}$3$}}
			& 13.02 & 6.99 & 0.024 & 22.48 & 2.07 & 0.009 \\
			\multicolumn{1}{l|}{\textsc{ChatGPT w/ \textbf{Tp}$4$}}
			& 13.29 & 7.92 & \textbf{0.005} & 11.97 & 3.65 & 0.006 \\
			\multicolumn{1}{l|}{\textsc{ChatGPT w/ \textbf{Tp}$5$}}
			& 12.00 & 14.36 & 0.027 & 20.48 & 7.74 & 0.020 \\
			\multicolumn{1}{l|}{\textsc{ChatGPT w/ \textbf{Tp}$6$}}
			& 9.68 & 13.09 & 0.028 & 17.24 & 8.40 & 0.040 \\
			
			\hline\hline
		\end{tabular}
	\end{center}
	\caption{\label{dup} Number and duplication ratio of predicted keyphrases on two keyphrase extraction datasets. “\#PK” and “\#AK” are the average number of unique present and absent keyphrases respectively. “Dup” refers to the average duplication ratio of predicted keyphrases. “Oracle” refers to the ground truth.}
\end{table}
\subsection{Keyphrase Generation Diversity}
%To investigate the ability of ChatGPT to generate diverse keyphrases, we measure the average numbers of unique present and absent keyphrases and the average duplication ratio of all the predicted keyphrases following recent studies \cite{one2set, wrone2set}. The results are reported in Table 4. Based on the results, ChatGPT generates more unique keyphrases than all baselines by a large margin and achieves a significantly lower duplication ratio without additional hyper-parameter settings.
To investigate the ability of ChatGPT to produce diverse keyphrases, we computed the average counts of unique present and absent keyphrases, as well as the average duplication ratio of all the generated keyphrases, in accordance with recent research, as documented in \cite{one2set} and \cite{wrone2set}. The results are presented in Table 4. According to the findings, ChatGPT generates a significantly higher number of unique keyphrases than all the baselines and achieves a significantly lower duplication ratio without additional hyper-parameter adjustments.

\section{Conclusion}
This report presents a preliminary study of ChatGPT for the keyphrase generation task, including keyphrase generation prompts, and keyphrase generation diversity.
The results from the Inspec dataset in Table~\ref{present} and Table~\ref{absent} show that ChatGPT has a strong keyphrase generation capability. It just needs to design better prompts to guide it. In the following section, we list several aspects not considered in this report and argue that these may affect the keyphrase generation performance of ChatGPT.

\section{Limitations}
We admit that this report is far from complete with various aspects to make it more reliable.
\subsection{Powerful Prompt Designing}
Generally, when chatting with ChatGPT, the design of prompts can largely influence the results it gives. In this report, we make some improvements based on the prompts given by OpenAI, but they are not necessarily optimal. Therefore, designing more appropriate prompts is the key to effectively exploiting the performance of ChatGPT on the task of keyphrase generation.

\subsection{Hyper-Parameter Settings}
In realistic scenes, users may not care about setting hyper-parameters of ChatGPT when chatting with ChatGPT. Meanwhile, the settings of the hyper-parameters typically require prior knowledge about ChatGPT. Therefore, we do not consider setting different hyper-parameters in this report, which affected the keyphrase generation performance of ChatGPT. Next, we will further consider the corresponding hyper-parameters of ChatGPT to verify its performance on the keyphrase generation task and give a more detailed analysis.

\subsection{Few-Shot Prompting}
With the increasing power of large language models, in-context learning has become a new paradigm for natural language processing \cite{icl}, where large language models make predictions only based on contexts augmented with a few examples. It has been a new trend to explore in-context learning to evaluate and extrapolate the ability of large language models (e.g., the emergent ability \cite{emergent}). In this report, we do not consider using some strategies to explore the ability of ChatGPT. In the future direction, we believe it is possible to enhance the performance of large language models in the keyphrase generation task through similar methods. 

\subsection{Supervised Fine-Tuning}
In this paper, the selected baselines are based on supervised learning; however, the results obtained by ChatGPT are achieved in a zero-shot setting. Therefore, achieving better results than existing supervised keyphrase generation methods should be possible by fine-tuning large language models through supervised learning.

\subsection{Evaluation Metric} 
Previous studies \cite{catseq17, one2set} have mainly used extensions of standard F1-based metrics to measure the performance of keyphrase generation models. Such evaluation metrics usually operate based on exact matches between predicted and gold keyphrases. Such a strategy cannot account for partial matches or semantic similarity. For example, if the prediction is "keyphrase generation model" and the gold is "keyphrase generation system", despite both semantic similarity and partial matching, the score will be 0. These minor deviations are ubiquitous in keyphrase generation yet harshly penalized by the "exact match" evaluation metrics. Therefore, a semantic-based evaluation metric may be more suitable to measure the performance of ChatGPT on the keyphrase generation task. Furthermore, human evaluation can provide more insights for comparing ChatGPT with keyphrase generation baselines.

\bibliography{anthology}

\begin{thebibliography}{27}
\expandafter\ifx\csname natexlab\endcsname\relax\def\natexlab#1{#1}\fi

\bibitem[{Bahuleyan and Asri(2020)}]{diversity}
Hareesh Bahuleyan and Layla~El Asri. 2020.
\newblock \href
  {http://dblp.uni-trier.de/db/journals/corr/corr2010.html#abs-2010-07665}
  {Diverse keyphrase generation with neural unlikelihood training.}
\newblock \emph{CoRR}, abs/2010.07665.

\bibitem[{Bang et~al.(2023)Bang, Cahyawijaya, Lee, Dai, Su, Wilie, Lovenia, Ji,
  Yu, Chung, Do, Xu, and Fung}]{gangda}
Yejin Bang, Samuel Cahyawijaya, Nayeon Lee, Wenliang Dai, Dan Su, Bryan Wilie,
  Holy Lovenia, Ziwei Ji, Tiezheng Yu, Willy Chung, Quyet~V. Do, Yan Xu, and
  Pascale Fung. 2023.
\newblock \href {https://doi.org/10.48550/arXiv.2302.04023} {A multitask,
  multilingual, multimodal evaluation of chatgpt on reasoning, hallucination,
  and interactivity}.
\newblock \emph{CoRR}, abs/2302.04023.

\bibitem[{Chan et~al.(2019)Chan, Chen, Wang, and King}]{adaptive_reward}
Hou~Pong Chan, Wang Chen, Lu~Wang, and Irwin King. 2019.
\newblock \href {https://doi.org/10.18653/v1/P19-1208} {Neural keyphrase
  generation via reinforcement learning with adaptive rewards}.
\newblock In \emph{Proceedings of the 57th Annual Meeting of the Association
  for Computational Linguistics}, pages 2163--2174, Florence, Italy.
  Association for Computational Linguistics.

\bibitem[{Chen et~al.(2020)Chen, Chan, Li, and King}]{ExHiRD}
Wang Chen, Hou~Pong Chan, Piji Li, and Irwin King. 2020.
\newblock \href {https://doi.org/10.18653/v1/2020.acl-main.103} {Exclusive
  hierarchical decoding for deep keyphrase generation}.
\newblock In \emph{Proceedings of the 58th Annual Meeting of the Association
  for Computational Linguistics}, pages 1095--1105, Online. Association for
  Computational Linguistics.

\bibitem[{Devlin et~al.(2019)Devlin, Chang, Lee, and Toutanova}]{bert}
Jacob Devlin, Ming-Wei Chang, Kenton Lee, and Kristina Toutanova. 2019.
\newblock \href
  {http://dblp.uni-trier.de/db/conf/naacl/naacl2019-1.html#DevlinCLT19} {Bert:
  Pre-training of deep bidirectional transformers for language understanding.}
\newblock In \emph{NAACL-HLT (1)}, pages 4171--4186. Association for
  Computational Linguistics.

\bibitem[{Dong et~al.(2022)Dong, Li, Dai, Zheng, Wu, Chang, Sun, Xu, and
  Sui}]{icl}
Qingxiu Dong, Lei Li, Damai Dai, Ce~Zheng, Zhiyong Wu, Baobao Chang, Xu~Sun,
  Jingjing Xu, and Zhifang Sui. 2022.
\newblock A survey for in-context learning.
\newblock \emph{arXiv preprint arXiv:2301.00234}.

\bibitem[{Hasan and Ng(2014)}]{2014survey}
Kazi~Saidul Hasan and Vincent Ng. 2014.
\newblock \href {http://dblp.uni-trier.de/db/conf/acl/acl2014-1.html#HasanN14}
  {Automatic keyphrase extraction: A survey of the state of the art.}
\newblock In \emph{ACL (1)}, pages 1262--1273. The Association for Computer
  Linguistics.

\bibitem[{Hulth(2003)}]{Inspec}
Anette Hulth. 2003.
\newblock \href {http://dblp.uni-trier.de/db/conf/emnlp/emnlp2003.html#Hulth03}
  {Improved automatic keyword extraction given more linguistic knowledge.}
\newblock In \emph{EMNLP}.

\bibitem[{Kim et~al.(2010)Kim, Medelyan, Kan, and Baldwin}]{SemEval}
Su~Nam Kim, Olena Medelyan, Min-Yen Kan, and Timothy Baldwin. 2010.
\newblock \href
  {http://dblp.uni-trier.de/db/conf/semeval/semeval2010.html#KimMKB10}
  {Semeval-2010 task 5 : Automatic keyphrase extraction from scientific
  articles.}
\newblock In \emph{SemEval@ACL}, pages 21--26. The Association for Computer
  Linguistics.

\bibitem[{Krapivin and Marchese(2009)}]{Krapivin}
M.~Krapivin and M.~Marchese. 2009.
\newblock Large dataset for keyphrase extraction.

\bibitem[{Meng et~al.(2017)Meng, Zhao, Han, He, Brusilovsky, and
  Chi}]{catseq17}
Rui Meng, Sanqiang Zhao, Shuguang Han, Daqing He, Peter Brusilovsky, and
  Yu~Chi. 2017.
\newblock \href
  {http://dblp.uni-trier.de/db/conf/acl/acl2017-1.html#MengZHHBC17} {Deep
  keyphrase generation.}
\newblock In \emph{ACL}, pages 582--592. Association for Computational
  Linguistics.

\bibitem[{Nguyen and Kan(2007)}]{Nus}
Thuy~Dung Nguyen and Min-Yen Kan. 2007.
\newblock \href
  {http://dblp.uni-trier.de/db/conf/icadl/icadl2007.html#NguyenK07} {Keyphrase
  extraction in scientific publications.}
\newblock In \emph{ICADL}, volume 4822 of \emph{Lecture Notes in Computer
  Science}, pages 317--326. Springer.

\bibitem[{Ouyang et~al.(2022)Ouyang, Wu, Jiang, Almeida, Wainwright, Mishkin,
  Zhang, Agarwal, Slama, Ray, Schulman, Hilton, Kelton, Miller, Simens, Askell,
  Welinder, Christiano, Leike, and Lowe}]{instructgpt}
Long Ouyang, Jeff Wu, Xu~Jiang, Diogo Almeida, Carroll~L. Wainwright, Pamela
  Mishkin, Chong Zhang, Sandhini Agarwal, Katarina Slama, Alex Ray, John
  Schulman, Jacob Hilton, Fraser Kelton, Luke Miller, Maddie Simens, Amanda
  Askell, Peter Welinder, Paul~F. Christiano, Jan Leike, and Ryan Lowe. 2022.
\newblock \href {https://doi.org/10.48550/arXiv.2203.02155} {Training language
  models to follow instructions with human feedback}.
\newblock \emph{CoRR}, abs/2203.02155.

\bibitem[{Song et~al.(2022{\natexlab{a}})Song, Feng, and Jing}]{hypermatch}
Mingyang Song, Yi~Feng, and Liping Jing. 2022{\natexlab{a}}.
\newblock \href {https://doi.org/10.18653/v1/2022.naacl-main.419} {Hyperbolic
  relevance matching for neural keyphrase extraction}.
\newblock In \emph{Proceedings of the 2022 Conference of the North American
  Chapter of the Association for Computational Linguistics: Human Language
  Technologies, {NAACL} 2022}, pages 5710--5720.

\bibitem[{Song et~al.(2022{\natexlab{b}})Song, Feng, and Jing}]{icnlsp}
Mingyang Song, Yi~Feng, and Liping Jing. 2022{\natexlab{b}}.
\newblock \href {https://aclanthology.org/2022.icnlsp-1.32} {Utilizing {BERT}
  intermediate layers for unsupervised keyphrase extraction}.
\newblock In \emph{5th International Conference on Natural Language and Speech
  Processing, {ICNLSP} 2022}, pages 277--281.

\bibitem[{Song et~al.(2023{\natexlab{a}})Song, Feng, and Jing}]{song_survey}
Mingyang Song, Yi~Feng, and Liping Jing. 2023{\natexlab{a}}.
\newblock \href {https://aclanthology.org/2023.findings-eacl.161} {A survey on
  recent advances in keyphrase extraction from pre-trained language models}.
\newblock In \emph{Findings of the Association for Computational Linguistics:
  {EACL} 2023}, pages 2108--2119. Association for Computational Linguistics.

\bibitem[{Song et~al.(2023{\natexlab{b}})Song, Jiang, Liu, Shi, and
  Jing}]{setmatch}
Mingyang Song, Haiyun Jiang, Lemao Liu, Shuming Shi, and Liping Jing.
  2023{\natexlab{b}}.
\newblock \href {https://aclanthology.org/2023.findings-acl.156} {Unsupervised
  keyphrase extraction by learning neural keyphrase set function}.
\newblock In \emph{Findings of the Association for Computational Linguistics:
  ACL 2023}, pages 2482--2494. Association for Computational Linguistics.

\bibitem[{Song et~al.(2021)Song, Jing, and Xiao}]{kiemp}
Mingyang Song, Liping Jing, and Lin Xiao. 2021.
\newblock \href {https://doi.org/10.18653/v1/2021.emnlp-main.215} {{I}mportance
  {E}stimation from {M}ultiple {P}erspectives for {K}eyphrase {E}xtraction}.
\newblock In \emph{Proceedings of the 2021 Conference on Empirical Methods in
  Natural Language Processing}, pages 2726--2736. Association for Computational
  Linguistics.

\bibitem[{Song et~al.(2023{\natexlab{c}})Song, Liu, Feng, and Jing}]{hguke}
Mingyang Song, Huafeng Liu, Yi~Feng, and Liping Jing. 2023{\natexlab{c}}.
\newblock \href {https://aclanthology.org/2023.findings-acl.66} {Improving
  embedding-based unsupervised keyphrase extraction by incorporating structural
  information}.
\newblock In \emph{Findings of the Association for Computational Linguistics:
  ACL 2023}, pages 1041--1048. Association for Computational Linguistics.

\bibitem[{Song et~al.(2023{\natexlab{d}})Song, Liu, and Jing}]{diversityrank}
Mingyang Song, Huafeng Liu, and Liping Jing. 2023{\natexlab{d}}.
\newblock \href {https://doi.org/10.1145/3583780.3615141} {Improving diversity
  in unsupervised keyphrase extraction with determinantal point process}.
\newblock In \emph{Proceedings of the 32nd ACM International Conference on
  Information and Knowledge Management}, CIKM '23, page 4294–4299.
  Association for Computing Machinery.

\bibitem[{Song et~al.(2023{\natexlab{e}})Song, Xiao, and Jing}]{csl}
Mingyang Song, Lin Xiao, and Liping Jing. 2023{\natexlab{e}}.
\newblock \href {https://doi.org/10.1016/j.csl.2023.101502} {Learning to
  extract from multiple perspectives for neural keyphrase extraction}.
\newblock \emph{Comput. Speech Lang.}, 81:101502.

\bibitem[{Sun et~al.(2021)Sun, Liu, Xiong, Liu, and Bao}]{baseline}
Si~Sun, Zhenghao Liu, Chenyan Xiong, Zhiyuan Liu, and Jie Bao. 2021.
\newblock Capturing global informativeness in open domain keyphrase extraction.
\newblock In \emph{CCF International Conference on Natural Language Processing
  and Chinese Computing}, pages 275--287. Springer.

\bibitem[{Wei et~al.(2022)Wei, Tay, Bommasani, Raffel, Zoph, Borgeaud,
  Yogatama, Bosma, Zhou, Metzler, Chi, Hashimoto, Vinyals, Liang, Dean, and
  Fedus}]{emergent}
Jason Wei, Yi~Tay, Rishi Bommasani, Colin Raffel, Barret Zoph, Sebastian
  Borgeaud, Dani Yogatama, Maarten Bosma, Denny Zhou, Donald Metzler, Ed~H.
  Chi, Tatsunori Hashimoto, Oriol Vinyals, Percy Liang, Jeff Dean, and William
  Fedus. 2022.
\newblock \href {https://doi.org/10.48550/arXiv.2206.07682} {Emergent abilities
  of large language models}.
\newblock \emph{CoRR}, abs/2206.07682.

\bibitem[{Wu et~al.(2022)Wu, Ahmad, and Chang}]{wudi}
Di~Wu, Wasi~Uddin Ahmad, and Kai-Wei Chang. 2022.
\newblock \href {https://doi.org/10.48550/ARXIV.2212.10233} {Pre-trained
  language models for keyphrase generation: A thorough empirical study}.

\bibitem[{Xie et~al.(2022)Xie, Wei, Yang, Lin, Xie, Wang, Zhang, and
  Su}]{wrone2set}
Binbin Xie, Xiangpeng Wei, Baosong Yang, Huan Lin, Jun Xie, Xiaoli Wang, Min
  Zhang, and Jinsong Su. 2022.
\newblock \href {https://aclanthology.org/2022.emnlp-main.491} {Wr-one2set:
  Towards well-calibrated keyphrase generation}.
\newblock In \emph{Proceedings of the 2022 Conference on Empirical Methods in
  Natural Language Processing, {EMNLP} 2022, Abu Dhabi, United Arab Emirates,
  December 7-11, 2022}, pages 7283--7293. Association for Computational
  Linguistics.

\bibitem[{Xiong et~al.(2019)Xiong, Hu, Xiong, Campos, and Overwijk}]{xiong19}
Lee Xiong, Chuan Hu, Chenyan Xiong, Daniel Campos, and Arnold Overwijk. 2019.
\newblock \href
  {http://dblp.uni-trier.de/db/conf/emnlp/emnlp2019-1.html#XiongHXCO19} {Open
  domain web keyphrase extraction beyond language modeling.}
\newblock In \emph{EMNLP/IJCNLP (1)}, pages 5174--5183. Association for
  Computational Linguistics.

\bibitem[{Ye et~al.(2021)Ye, Gui, Luo, Xu, and Zhang}]{one2set}
Jiacheng Ye, Tao Gui, Yichao Luo, Yige Xu, and Qi~Zhang. 2021.
\newblock \href {https://doi.org/10.18653/v1/2021.acl-long.354} {One2set:
  Generating diverse keyphrases as a set}.
\newblock In \emph{Proceedings of the 59th Annual Meeting of the Association
  for Computational Linguistics and the 11th International Joint Conference on
  Natural Language Processing, {ACL/IJCNLP} 2021, (Volume 1: Long Papers),
  Virtual Event, August 1-6, 2021}, pages 4598--4608. Association for
  Computational Linguistics.

\end{thebibliography}
\bibliographystyle{acl_natbib}
\end{document}